\documentclass[a4paper]{article}

\usepackage{INTERSPEECH2020}
\usepackage{graphicx}
\usepackage{ulem}
\usepackage{amsmath}
\usepackage{enumitem,kantlipsum}

\title{Comparing Natural Language Processing Techniques for Alzheimer's Dementia Prediction in Spontaneous Speech}
\name{Thomas Searle$^1$, Zina Ibrahim$^1$, Richard Dobson$^{1,2}$}
\address{
  $^1$Department of Biostatistics and Health Informatics,\\
  Institute of Psychiatry, Psychology and Neuroscience,\\
  King’s College London, London, U.K.\\
  $^2$Institute of Health Informatics, University College London,\\
  London, London, U.K.}
\email{\{firstname\}.\{lastname\}@kcl.ac.uk}

\begin{document}

\maketitle

\begin{abstract}
Alzheimer's Dementia (AD) is an incurable, debilitating, and progressive neurodegenerative condition that affects cognitive function. Early diagnosis is important as therapeutics can delay progression and give those diagnosed vital time. Developing models that analyse spontaneous speech could eventually provide an efficient diagnostic modality for earlier diagnosis of AD. The Alzheimer's Dementia Recognition through Spontaneous Speech task offers acoustically pre-processed and balanced datasets for the classification and prediction of AD and associated phenotypes through the modelling of spontaneous speech. We exclusively analyse the supplied textual transcripts of the spontaneous speech dataset, building and comparing performance across numerous models for the classification of AD vs controls and the prediction of Mental Mini State Exam scores. We rigorously train and evaluate Support Vector Machines (SVMs), Gradient Boosting Decision Trees (GBDT), and Conditional Random Fields (CRFs) alongside deep learning Transformer based models. We find our top performing models to be a simple Term Frequency-Inverse Document Frequency (TF-IDF) vectoriser as input into a SVM model and a pre-trained Transformer based model `DistilBERT' when used as an embedding layer into simple linear models. We demonstrate test set scores of 0.81-0.82 across classification metrics and a RMSE of 4.58.

\end{abstract}

\section{Introduction}
Alzheimer's Dementia (AD) is a progressive neurodegenerative condition that largely affects cognitive function. With our globally aging population, conditions such as AD are likely to become more prevalent\cite{Mayeux2012-ls}. Despite there being no cure currently, early diagnosis can offer interventions to slow or delay progression of symptoms\cite{Rasmussen2019-cb}. Prior work has used machine learning methods for the prediction of cognitive impairment (CI) conditions, including AD, using patient structured data\cite{Kang2019-yp} and medical imaging data\cite{Stonnington2010-fz}. Linguistic phenomenon have also been identified in those already diagnosed with AD\cite{Guo2019-zk,Zhou2016-jt}. 

The Alzheimer's Dementia Recognition through Spontaneous Speech (ADReSS) challenge presents two tasks in the modelling of spontaneous speech\cite{Luz2020-tn}. Firstly, to classify presence of AD vs controls and secondly, to predict the `Mental Mini State Exam' score, a common set of questions designed to assess cognitive function\cite{De_Boer2014-zs}. The challenge provides 108 training, 54 AD vs 54 Control samples, and 48 unseen test samples. Spontaneous speech audio and associated transcripts of participants describing the `Cookie Theft' picture from the Boston Diagnostic Aphasia Exam\cite{Goodglass2001-vm} are provided. Samples are demographically and acoustically balanced and longer in duration than previous clinical studies\cite{Luz2020-tn}. The challenge provides an environment for researchers to test competing methods with recommendations for future work. Using machine learning techniques to predict AD from spontaneous speech could potentially offer an efficient early diagnostic modality. For example, audio samples could be collected via a mobile device with results directing individuals to seek more formal medical evaluation.

\section{Data Prepossessing}\label{sec:data}
In this work we exclusively focus on the textual transcriptions that are provided alongside the audio samples. Transcripts are supplied using the CHAT transcription format\cite{MacWhinney2017-av}. The transcription schema provides the linguistic content alongside some prosodic content such as: pauses, laughter, discourse markers such as `um' and `ah', and abbreviations such as `(be)cause'. We preprocess each transcript before feeding into our model pipelines. All code to re-create the data prepossessing, experiments and analysis is available open-source\footnote{https://github.com/tomolopolis/ADReSS\_Challenge}. 

The preprocessing parses participant metadata such as age, sex, AD diagnosis and MMSE score. Each transcription line is parsed to remove time duration suffixes, specific speech artifacts such as `[',`]' or `\textgreater', `\textless' and excess white-space such as tabs and newlines. We purposely leave discourse markers such as `um' `ah' and other speech artifacts such as `+...', `\&=laughs' and `(...)' that indicate various pause types, or laughter in the audio. 

\subsection{Data Splits and Granularity}\label{sec:data_splits}
We split the transcripts into multiple competing datasets providing the candidate models with greatest opportunity to find adequate signal for the prediction tasks.

\subsubsection{Transcript Level Data}\label{data:coarse}
Within these datasets each transcript is a single data point with their corresponding AD label and assigned MMSE score. This includes:
\begin{enumerate}[leftmargin=*]
    \item A dataset with only participant utterances concatenated together into a single paragraph as they appear in the transcript. Denoted \textbf{PAR}.
    \item A dataset with both participant and interviewer speech concatenated into a single paragraph as they appear in the transcript. Denoted \textbf{PAR+INV}.
\end{enumerate}

\subsubsection{Utterance Level Data}\label{data:granular}
For these datasets we define each utterance as an individual data point. This provides N=1,476, AD(N=740), controls (N=736). The target labels and MMSE scores are replicated to each utterance. Segments maintain a reference to their source transcript so random shuffling does not produce data leakage between the train and test phases. We only consider participant spoken utterances here as initial experiments indicated including interviewer speech lead to a reduction in performance. This includes:
\begin{enumerate}[leftmargin=*]
    \item A dataset with only participant utterance as individual classification \& regression data points. Denoted \textbf{PAR\_SPLT}.
    \item Further datasets that extend the text based features with the inclusion of  temporal and participant demographic features such as: time duration per sentence, time between sentences, average/max/min sentence time denoted \textbf{PAR\_SPLT+T}, and participant age and sex denoted \textbf{PAR\_SPLT+T+D}.
\end{enumerate}

\section{Methods}
The baseline paper accompanying the challenge\cite{Luz2020-tn} only creates a single baseline result using only the text transcripts. Therefore, we present a range of models both as a new baseline result for the linguistic features, Section \ref{sec:meth_base},  alongside our more advanced approaches in Section \ref{sec:meth_dl}.

\subsection{Baseline Methods}\label{sec:meth_base}
We make extensive use of Scikit-Learn\cite{Pedregosa2011-wy}, a python based machine learning framework that provides APIs for common machine learning models, feature extraction, cross validation, hyper parameter optimisation and performance metric calculation. We use the integrated Term-Frequency-Inverse Document Frequency (TF-IDF)\cite{Luhn1957-ow} `bag-of-words' vectoriser. With this method text inputs forgo their sequence order and words are counted within and across documents. TF-IDF down-weights the counts of common cross-document terms, and increases weights of rare cross-document but frequent intra-document terms. This embedding method is a common first stage in any textual modelling exercise due to its efficiency and ease of use. 

Scikit-learn provides APIs for optimised implementations of common machine learning algorithms such as libsvm\cite{Chang2011-qa} for Support Vector Machines(SVM)\cite{Cortes1995-px} and XGBoost\cite{Chen2016-sr} for Gradient Boosted Decision Trees(GBDT)\cite{Friedman2001-yt} allowing for fast model fitting. We use both algorithms in the development of our baseline models for the transcript level and utterance level datasets presented in Section \ref{sec:data_splits}

SVMs and GBDTs are effective techniques to learn non-linear relationships between input features and the decision boundaries for both classification and regression tasks.

\subsubsection{Utterance Level Methods}
For the segmented speech datasets, PAR\_SPLT, PAR\_SPLT+T, PAR\_SPLT+T+D presented in Section \ref{data:granular}, our modelling approach does not support MMSE prediction so we only report results for AD classification. We train and cross validate TF-IDF/SVM and TF-IDF/GBDT models on each utterance, and feed output prediction probability sequences to a Conditional Random Field (CRF)\cite{Lafferty2001-ej}. CRFs are effective in the modelling of sequential data as input feature representations can depend on previous and future states of the sequence. For the overall classification of the transcript we take the final classification state of the CRF. 

\subsubsection{Hyper Parameter Optimisation}
Table \ref{tab:hyper_params} lists the model configuration and associated hyper-parameter spaces we search across during an exhaustive 5-fold cross validation grid-search. As our dataset is fairly small, performing this only took a couple of minutes for each model configuration and each dataset despite the many individual model fits.

\setlength\tabcolsep{1pt}
\begin{table}[th]
  \caption{Baseline methods hyper-parameter searched and found optimal parameters. $^{\ast}$ values are $\times10^3$. $^{\dagger{}}$ the parameter spaces are sampled from an exponential probability distributions 15 times with specified $\lambda$}
  \label{tab:hyper_params}
  \centering
  \begin{tabular}{ c c c c }
    \toprule
     Model &    Hyper Parameter         &   Param Space     & Optimal \\
    \midrule
     TF-IDF/GBDT & Max Features    & 0.1, 0.5, 1, 2, 10$\ast$& 1$\ast$ \\
     TF-IDF/GBDT & Stop Words      & english, None           & english        \\
     TF-IDF/GBDT & Analyser        & word, char              & word          \\
     TF-IDF/GBDT & sublinear TF    & True, False             & True          \\
     TF-IDF/GBDT & N-Estimators    & 100, 200, 500           & 100           \\
     TF-IDF/GBDT & Max Depth       & 3, 5, 10                & 5             \\
     TF-IDF/SVM & Max Features     & 0.1, 0.5, 1, 2, 10$\ast$    & 0.1$\ast$\\
     TF-IDF/SVM & Stop Words       & english, None          & None         \\
     TF-IDF/SVM & Analyser         & word, char             & word         \\
     TF-IDF/SVM & sublinear TF     & True, False            & True         \\
     TF-IDF/SVM & Kernel           & rbf, sigmoid           & sigmoid      \\
     TF-IDF/SVM & C                & 0.1, 0.5, 1           & 1          \\
     SVM+CRF & c1               & $\lambda = 0.5^{\dagger{}}$   & 0.0036 \\
     SVM+CRF & c2               & $\lambda = 0.05^{\dagger{}}$  & 0.018  \\
     GBDT+CRF & c1               & $\lambda = 0.5^{\dagger{}}$  & 0.314   \\
     GBDT+CRF & c2               & $\lambda = 0.05^{\dagger{}}$  & 0.009  \\
    \bottomrule
  \end{tabular}
\end{table}

\subsection{Deep Learning Methods}\label{sec:meth_dl}
To converge successfully deep learning (DL) models often require more training data than methods such as SVMs and GBDTs. Training set sizes are often 50 or 100 times larger than  available here. Transfer learning presents a compelling option to enable re-use of deep learning models for smaller domain specific data sets. Recently, transfer learning approaches have been successfully applied to a variety of NLP problems\cite{Ruder2019-ks}.

Large pre-trained language models are an example of transfer learning, and can be used to provide semantically rich embedding layers, allowing researchers to re-use knowledge acquired by the model from a prior training process. The language modelling task can be defined as predicting the next word given the sequence of previous words, or formally in Equation \ref{eq:lm}, modelling the probability distribution of all words $w$ in a vocabulary $V$ conditioned on previous words $w_{i-1}$ to $w_1$. 

\begin{equation}\label{eq:lm}
  P(w_i | w_{i-1}, w_{i-2} \cdots w_1) \forall w \in V 
\end{equation}

The task enables the usage of large corpora of existing texts without any explicit manual annotation, often referred to as self-supervised learning\cite{Mikolov2013-ib}. Each model we use is based upon the Transformer architecture first presented for sequence to sequence problems such as machine translation\cite{Vaswani2017-xe}. The Transformer consists of layers of encoder and decoder blocks of multi-headed self-attention followed by fully connected layers. Each successive layer learns sophisticated latent representations of the input texts.

We use the `transformers'\cite{Wolf2019-mo} library to load, and re-use the BERT\cite{Devlin2018-mq}, RoBERTa\cite{Liu2019-wm} and DistilBERT/DistilRoBERTa\cite{Sanh2019-hj} models as embedding layers for the PAR and PAR+INV datasets. 

Running the input transcripts through the pre-trained models produces a fixed size embedding representation for each provided transcript. This is an embedding matrix of size $N \times H$, where $N$ is the number of transcripts and $H$ is the hidden dimension of the pre-trained model. As recommended in prior work we fit simple linear models, Logistic Regression model for AD classification and LASSO Regression for MMSE prediction, to produce our final predictions.

\section{Results}\label{sec:res}
Table \ref{tab:results} provides results for average 10 fold cross-validation for hyper parameter selection and best train/development set performance. This attempts to compare model robustness with the available training data, especially for our transcript level datasets where dataset size is limited. Metrics follow the standard definitions as outlined in the baseline work\cite{Luz2020-tn} and are averaged between the classes Non-AD / AD for precision, recall and F1. We then pick our 5 best performing models / dataset configurations and run on the unlabelled test dataset, containing 48 samples, sending our AD and MMSE predictions to challenge organisers. Organisers subsequently responded with aggregate results as reported in Section \ref{res:ad} for AD classification and Section \ref{res:mmse} for MMSE predictions.

\setlength\tabcolsep{1pt} 
\begin{table}[th]
  \caption{Average 10-fold CV AD Classification and MMSE prediction results. Results are highlighted if within 0.02 of the highest score. $^{\ast}$ indicates best score for given metric.}
  \label{tab:results}
  \centering
  \begin{tabular}{ c c c c c c c }
    \toprule
     Dataset &  Model         &  Acc     & Prec & Recall & F1 & RMSE\\
    \midrule
     PAR     &  GBDT          &   .82     & .84 & .82 & .81 & 5.93\\
     PAR     &  SVM           &   .86     & \textbf{.90} & .83 & \textbf{.86} & 6.57\\
     PAR     & DistilBERT     &   \textbf{.87} & \textbf{.90} & .87 & \textbf{.87} & \textbf{4.49$^{\ast}$}\\
     PAR     & DistilRoBERTa  &   .84     & .86 & .85 & .82 & 5.12\\
     PAR     & BERT(base)     &   .84     & .86 & .85 & .82 & 5.12\\
     PAR     & RoBERTa(base)  &   .75     & .79 & .72 & .74 & 7.11\\
     PAR     & BERT(large)    &   .77     & .80 & .77 & .76 & 6.64\\
     PAR     & RoBERTa(large) &   .77     & .81 & .73 & .76 & 7.13\\
     PAR+INV &  GBDT          &   .79     & .80 & .82 & .79 & 5.60\\
     PAR+INV &  SVM           &   \textbf{.88} & \textbf{.92$^{\ast}$} &.87 & \textbf{.87} & 6.74\\
     PAR+INV & DistilBERT     & \textbf{.87}& .89 & \textbf{.89} & \textbf{.88$^{\ast}$} & 4.85\\
     PAR+INV & DistilRoBERTa  &   .80     & .87 & .79 & .78 & 7.11\\
     PAR+INV & BERT(base)     &   .75     & .76 & .78 & .74 & 7.13\\
     PAR+INV & RoBERTa(base)  &   .72     & .71 & .71 & .69 & 5.45\\
     PAR+INV & BERT(large)    &   .75     & .78 & .73 & .74 & 7.13\\
     PAR+INV & RoBERTa(large) &   .81     & .88 & .76 & .79 & 6.64\\
     PAR\_SPLT & SVM+CRF       &   .88     & .88 & \textbf{.88} & \textbf{.87} & - \\
     PAR\_SPLT & GBDT+CRF      &   .80     & .84 & .74 & .78 & - \\
     PAR\_SPLT+T & SVM+CRF     &   \textbf{.89$^{\ast}$} & .87 & \textbf{.90$^{\ast}$} & \textbf{.88$^{\ast}$} & - \\
     PAR\_SPLT+T & GBDT+CRF    &   .82     & .84 & .79 & .81 & - \\
     PAR\_SPLT+T+D & SVM+CRF     & .86     & .85 & .87 & \textbf{.86} & - \\
     PAR\_SPLT+T+D & GBDT+CRF    & .83     & .86 & .79 & .81 & - \\
    \bottomrule
  \end{tabular}
\end{table}

\subsection{AD Classification}\label{res:ad}
Table \ref{tab:test_res_ad} shows our test set results for each metric. We show results for metrics both labels (AD vs No AD) for precision, recall and F1 metrics as defined in baseline work\cite{Luz2020-tn}. 

\setlength\tabcolsep{1.5pt}
\begin{table}[th]
    \caption{Test set results for AD classification}
    \centering
    \label{tab:test_res_ad}
    \begin{tabular}{c c c c c c }
         \toprule
            Dataset / Model & Class & Prec & Recall & F1 & Acc\\
        \midrule 
                PAR / DistilBERT & 
                \begin{tabular}{@{}c@{}}Non-AD \\ AD\end{tabular} &
                \begin{tabular}{@{}c@{}}0.76 \\ 0.783\end{tabular} & 
                \begin{tabular}{@{}c@{}}0.79 \\ 0.75\end{tabular} & 
                \begin{tabular}{@{}c@{}}0.78 \\ 0.77\end{tabular} &
                0.77 \\
            \hline
                PAR+INV / DistilBERT & 
                \begin{tabular}{@{}c@{}}Non-AD \\ AD\end{tabular} &
                \begin{tabular}{@{}c@{}}\textbf{0.83} \\ 0.80\end{tabular} & 
                \begin{tabular}{@{}c@{}}0.79 \\ \textbf{0.83}\end{tabular} & 
                \begin{tabular}{@{}c@{}}0.81 \\ \textbf{0.82}\end{tabular} &
                \textbf{0.81}  \\
            \hline
                PAR / TF-IDF/SVM &  
                \begin{tabular}{@{}c@{}}Non-AD \\ AD\end{tabular} &
                \begin{tabular}{@{}c@{}}0.70 \\ 0.79\end{tabular} & 
                \begin{tabular}{@{}c@{}}0.83 \\ 0.63\end{tabular} & 
                \begin{tabular}{@{}c@{}}0.75 \\ 0.70\end{tabular} & 
                0.73 \\
            \hline
                PAR\_SPLT / SVM+CRF & 
                \begin{tabular}{@{}c@{}}Non-AD \\ AD\end{tabular} &
                \begin{tabular}{@{}c@{}}0.78 \\ \textbf{0.86}\end{tabular} & 
                \begin{tabular}{@{}c@{}}0.88 \\ 0.75\end{tabular} & 
                \begin{tabular}{@{}c@{}}\textbf{0.82} \\ 0.80\end{tabular} & 
                \textbf{0.81}\\
            \hline
                PAR\_SPLT+T / SVM+CRF & 
                \begin{tabular}{@{}c@{}}Non-AD \\ AD\end{tabular} &
                \begin{tabular}{@{}c@{}}0.75 \\ 0.85\end{tabular} & 
                \begin{tabular}{@{}c@{}}\textbf{0.88} \\ 0.71\end{tabular} & 
                \begin{tabular}{@{}c@{}}0.81 \\ 0.77\end{tabular} & 
                0.79\\
            \hline
        \bottomrule
    \end{tabular}
\end{table}

\subsection{MMSE Prediction}\label{res:mmse}
Table \ref{tab:test_res_mmse} provides our MMSE prediction results on the provided test set. We observe that the deep learning embedding methods perform best, and in particular the DistilBERT model using only participant sections of the transcript performs best RMSE. Interestingly, the deep learning methods perform well despite having not been trained with regression tasks in mind. Our CRF models do not support regression so we cannot report MMSE prediction scores for those model configurations. 

\setlength\tabcolsep{3pt}
\begin{table}[th]
    \caption{Test set results for MMSE score prediction, `DBL' indicates our DistilBERT embedding with LASSO linear model.}
    \centering
    \label{tab:test_res_mmse}
    \begin{tabular}{c c c c}
         \toprule
            Dataset/Model & PAR/DBL & PAR+INV/DBL & PAR/SVM \\ 
            RMSE Score &    5.37        &       4.58 & 5.22\\
        \bottomrule
    \end{tabular}
\end{table}



\subsection{Alternative Configurations}\label{sec:alt_config}
The supplied transcripts included temporal metadata for each sentence for participants and interviewers. We experimented with these time time based features alongside the text features at the transcript level, i.e. a PAR+TIME dataset. This included features: participant average / minimum / maximum and median sentence times, and time between sentences. Intuitively, we assumed that AD subjects would exhibit distinctly different time based features due to their impaired cognitive function. However, this dataset (PAR+TIME) performed poorly across the modelling approaches so we do not include in our results. We suggest this is due to the aggregation in the transcript level dataset removing any signal to that could be detected by the modelling approaches. PAR\_SPLT+T does include temporal level features but does not perform as as well as linguistic features only.

\section{Discussion}
We discuss our results in context of model complexity, model generalisability and potential utility as a diagnostic modality. Our most effective models are DistilBERT with PAR+INV and SVM+CRF with PAR\_SPLT. Both models perform similarly for the AD classification task, but the deep learning approach can also output MMSE score predictions. The DL methods will likely generalise better as the majority of the modelling is accomplished by the embedding layer. Both models could be deployed to mobile devices for a potentially ubiquitous early diagnostic tool.  

In a potential diagnostic scenario, models would seek to balance recall and precision. A true-positive label of AD would prompt the user to seek a formal evaluation, under medical supervision, potentially leading to an earlier diagnosis allowing for slower progression of the disorder. However, ensuring the false-positive rate is low would minimise unnecessary anxiety during the following formal clinical evaluation.

\subsection{Baseline Approaches}
The SVM models report higher performance than the GBDT models across all metrics and tasks. They are also computationally faster to fit and cross validate. It is unclear if these models will generalise to alternative or larger datasets. The models have captured correlations in frequency of appearances of `key words' as identified by TF-IDF vectoriser. Further datasets may result in variations in performance as the frequencies of 'key words' change providing insufficient signal for accurate modelling of the decision boundaries necessary for prediction.

Despite offering the worst performance across all configurations and datasets, GBDTs do provide good model interpretability. For example, we find the top 20 most informative word level features from the TF-IDF vectoriser contain discourse markers such as `oh', `uh' and `um' for both tasks. This indicates the model has found occurrences of these words are useful in making predictions for both tasks. However, prior work has suggested more informative features are more complex\cite{Yancheva2015-qz}. 

\subsection{Deep Learning Approaches}
Our results are inline with previous work that has empirically shown that large, pre-trained, DL Transformer based, language models are an effective embedding layer that capture a variety of linguistic phenomenon\cite{Jawahar2019-ya}. As models are pre-trained we incur no model fit expense to use them. Training from scratch requires days or weeks with specialised hardware and large data sets. The simple LR or LASSO models that are fit on top of the fixed size output embeddings are as efficient to fit as the baseline SVM models.

BERT and RoBERTa models are available in their `base' and `large' varieties. In prior work, `large' often performs better due to the increased parameter space and longer training time\cite{Devlin2018-mq}. However, we observe in our experiments `large' models are often equivalent or worse performing. We also observe this trend with DistilBERT / DistilRoBERTa that have further reduced parameters compared to `base' varieties that broadly produce better results in our experiments although prior work would suggest the contrary.

\section{Future Work}

\subsection{Further NLP Modelling}
For future work we would look to replicate findings with larger datasets to demonstrate model robustness. We also currently use the models `out-of-the-box' so they have only been trained with large corpora of prepared speech (Wikipedia and the Toronto Book Corpus\cite{Devlin2018-mq}). For future work we would also look to fine-tune the deep learning embedding models specifically to spontaneous speech as fine-tuning to domain specific data often boosts performance\cite{Howard2018-fq}. Spontaneous speech corpora would likely show a difference in lexicon and grammar as well subtle prosodic differences such as rhythm, tempo that are often captured within spontaneous speech transcripts. An example of such a corpus is the `The British National Corpus\cite{Consortium2007-ux}'. A large corpora of informal spontaneous speech containing 1251 recordings and $\sim$11 million words from 668 speakers. We have cleaned and prepared the corpus using a sliding sentence window producing a dataset of $\sim$767k `documents'. We successfully begun the fine-tuning process observing a reduction in training loss. However, due to extenuating circumstances our GPU resource became unavailable and we were unable to complete the fine-tuning. We make the data pre-processing, and language model fine-tuning scripts available open-source\footnote{https://github.com/tomolopolis/ADReSS\_Challenge/blob/master/Fine-Tune-LanguageModel.ipynb}. 

\subsection{Feature Combinations and Model Ensembling}
Combining features or model ensembling that incorporated the acoustic data (i.e. prosodic/articulatory features) may provide further gains in performance. Audible phenomenon such as changes in pitch, intonation, stress and subtle changes in tempo would only be available in the audio dataset and have shown to be useful during prior work\cite{Konig2015-yf,Yancheva2015-qz}. We leave the investigation and ensembling of these features to future work.

\section{Conclusions}
We have presented a range of NLP techniques applied to the ADReSS challenge dataset, a shared task for the prediction of AD and MMSE scores of AD patients and controls. Each dataset and model configuration is rigorously optimised and tested. We observe promising results, above published baselines, for machine learning techniques such as SVMs and Deep Learning approaches. We highlight that the Deep Learning approaches are particularly effective when used as embedding layers for both the AD classification and MMSE score prediction tasks even despite the lack of domain and task specific fine-tuning.

\section{Acknowledgements}
RD's work is supported by 1.National Institute for Health Research (NIHR) Biomedical Research Centre at South London and Maudsley NHS Foundation Trust and King’s College London. 2. Health Data Research UK, which is funded by the UK Medical Research Council, Engineering and Physical Sciences Research Council, Economic and Social Research Council, Department of Health and Social Care (England), Chief Scientist Office of the Scottish Government Health and Social Care Directorates, Health and Social Care Research and Development Division (Welsh Government), Public Health Agency (Northern Ireland), British Heart Foundation and Wellcome Trust. 3. The National Institute for Health Research University College London Hospitals Biomedical Research Centre. This paper represents independent research part funded by the National Institute for Health Research (NIHR) Biomedical Research Centre at South London and Maudsley NHS Foundation Trust and King’s College London. The views expressed are those of the author(s) and not necessarily those of the NHS, MRC, NIHR or the Department of Health and Social Care.

\bibliographystyle{IEEEtran}

\bibliography{template}

\end{document}